\documentclass{article}

\usepackage{amsmath}
\usepackage{amsfonts}
\usepackage{times}
\usepackage{graphicx}
\usepackage{subfigure} 
\usepackage{natbib}
\usepackage{algorithm}
\usepackage[noend]{algorithmic}
\usepackage{amsmath}
\usepackage{hyperref}
\usepackage{mathtools}
\usepackage{icmlModified} 

\newtheorem{defn}{Definition}

\makeatletter
\newcommand{\ntimes}[2][\times]{
  \def\nextitem{\def\nextitem{#1}}
  \@for \el:=#2\do{\nextitem\el}
}
\newcommand{\R}[1]{{\rm I\!R}^{\ntimes{#1}}}
\makeatother

\runningtitle{A Manifold Approach to Learning Mutually Orthogonal Subspaces}

\begin{document} 

\twocolumn[
\title{A Manifold Approach to Learning Mutually Orthogonal Subspaces}
\author{Stephen Giguere}{sgiguere@cs.umass.edu}
\author{Francisco Garcia}{fmgarcia@cs.umass.edu}
\address{College of Information and Computer Sciences, University of Massachusetts, Amherst, Massachusetts 01003}
\author{Sridhar Mahadevan}{sridhar.mahadevan@sri.com}
\address{Stanford Research Institute, Menlo Park, California 94025}
\vskip 0.3in
]
\pagenumbering{gobble}

\begin{abstract}
    Although many machine learning algorithms involve learning subspaces with particular characteristics, optimizing a parameter matrix that is constrained to represent a subspace can be challenging.
    One solution is to use \emph{Riemannian optimization methods} that enforce such constraints implicitly, leveraging the fact that the feasible parameter values form a manifold.
    While Riemannian methods exist for some specific problems, such as learning a single subspace, there are more general subspace constraints that offer additional flexibility when setting up an optimization problem but have not been formulated as a manifold.

    We propose the \emph{partitioned subspace (PS) manifold} for optimizing matrices that are constrained to represent one or more subspaces.
    Each point on the manifold defines a partitioning of the input space into mutually orthogonal subspaces, where the number of partitions and their sizes are defined by the user.
    As a result, distinct groups of features can be learned by defining different objective functions for each partition.
    We illustrate the properties of the manifold through experiments on multiple dataset analysis and domain adaptation.
\end{abstract} 

\section{Introduction}
\pagenumbering{arabic}
\label{sec:introduction}

The process of designing a model and learning its parameters by numerically optimizing a loss function is a cornerstone of machine learning.
In this setting, it is common to place constraints on the model's parameters to ensure that they are valid, to promote desirable characteristics such as sparsity, to incorporate domain knowledge, or to make learning more efficient.
Because they can significantly impact the difficulty of an optimization problem, it is useful to understand the properties of particular types of constraints and to develop optimization techniques that preserve them.

Recently, there has been interest in developing \emph{Riemannian optimization} methods that enforce certain constraints by leveraging the geometry of the parameters that satisfy them, called the \emph{feasible set}.
Specifically, if the feasible set forms a smooth manifold in the original parameter space, then Riemannian optimization can be applied.
Unlike other strategies for enforcing constraints, such as augmenting the loss function with penalty terms or projecting the parameters back onto the feasible set, this approach eliminates the need to deal with constraints explictly by performing optimization on the constraint manifold directly.
In many cases, using Riemannian optimization can simplify algorithms, provide convergence guarantees on learning, and ensure that constraints are satisfied exactly rather than approximately.

\begin{figure*}[t]
    \centering
    \includegraphics[scale=0.85]{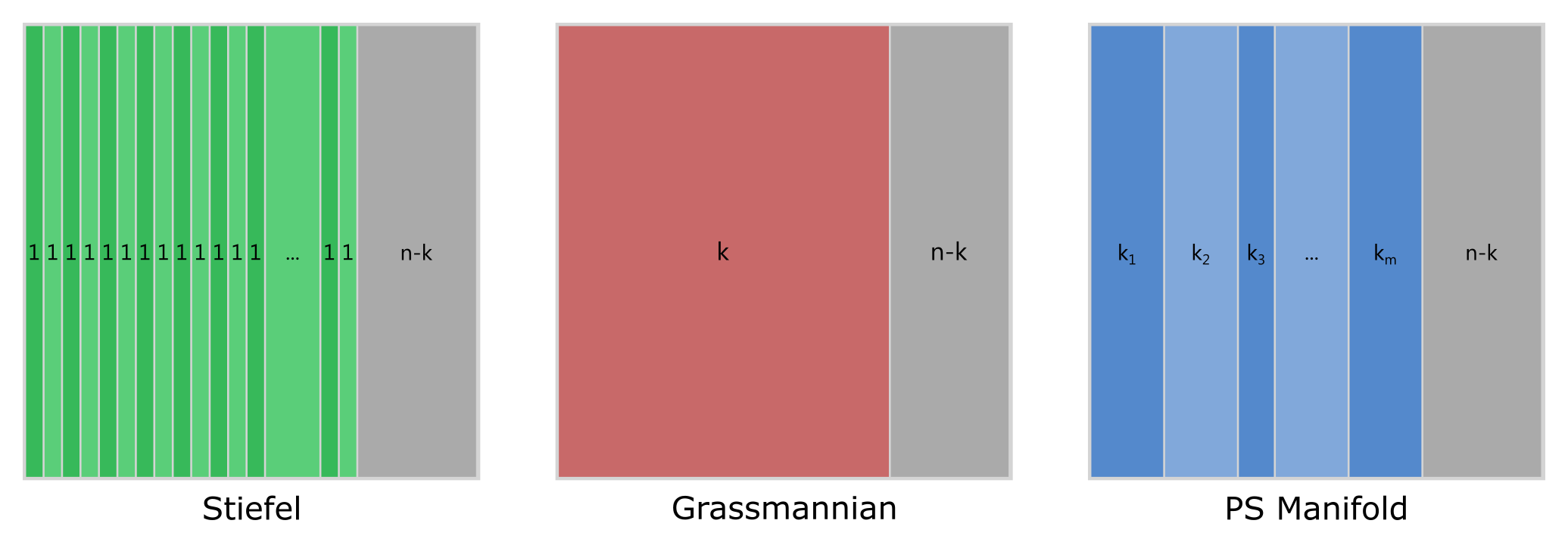}
    \caption{Visualization of $Q$ matrices showing the subspaces encoded by the Stiefel, Grassmannian, and partitioned subspace manifolds. Within each matrix, the colored segments represent mutually orthogonal subspaces. The values in each segment denote the dimensionality of the subspace. All three matrices are the same size and have $k$ columns of interest, but the properties of their respective manifolds cause the subspaces to be partitioned differently. While the Stiefel manifold represents $k$, $1$-dimensional subspaces and the Grassmannian represents a single, $k$-dimensional subspace, the PS manifold allows the number of partitions and their sizes to be chosen as needed for a given problem.}
    \label{fig:manifold_qs}
\end{figure*}

In this work, we investigate Riemannian optimization methods for enforcing \emph{subspace constraints}, which we define as any constraint that forces a matrix of parameters to represent one or more subspaces.
There are two commonly used subspace constraints that have well-established Riemannian optimization methods provided in \cite{edelman1998geometry}.
The first is applicable when optimizing over matrices that define $k$-dimensional bases in an $n$-dimensional ambient space.
In this case, the feasible set consists of all $n \times k$ matrices $Y$ that satisfy $Y^TY = I$, which corresponds to the \emph{Stiefel manifold}.
If the optimization is instead taken over all distinct $k$-dimensional subspaces, then in addition to the constraint that $Y^TY=I$, two matrices must be considered identical if they have the same span. 
This second condition is important because it implies that during optimization, estimates must be updated to not only change the parameter matrix, but to do so in a way that changes the subspace it represents.
The feasible set for this constraint corresponds to the \emph{Grassmannian manifold}, which has proven useful for a large number of applications including background separation in video, human activity analysis, subspace tracking, and others \cite{he2011online,turaga2009locally,he2012incremental}.

While the constraints satisfied by optimizing on the Stiefel or Grassmannian manifolds are useful in practice, there are more general subspace constraints that are not captured by these manifolds and offer significant flexibility when setting up an optimization problem.
To see this, first consider that a point on a Stiefel manifold can be viewed as a collection of $k$, $1$-dimensional subspaces that are constrained to be mutually orthogonal.
Similarly, a point on a Grassmannian manifold represents $1$, $k$-dimensional subspace.
It is therefore natural to consider a general constraint in which both the number of subspaces and their sizes can be specified by the user.
These relationships are shown in figure \ref{fig:manifold_qs}.
Importantly, subspace constraints of this form allow different set of features to be learned according to different criteria.
In analogy to optimization on the Stiefel and Grassmannian manifolds, we aim to optimize over distinct sets of $m$ mutually orthogonal subspaces with sizes $[k_1, k_2, ..., k_m]$, which we refer to as \emph{partitions}.

In this paper, we introduce a novel manifold that generalizes the Stiefel and Grassmannian manifolds to implicitly enforce the general subspace constraints described above.
Individual points on our proposed manifold define partitions of $n$-dimensional space into mutually orthogonal subspaces of particular sizes.
In addition, we derive update rules for performing Riemannian optimization on the manifold.
This allows features of the original space to be grouped in useful ways by defining separate objective functions on each subspace.
For example, given two datasets $X_0$ and $X_1$, the linear features that best describe the two can easily be partitioned into a pair of subspaces containing features unique to each dataset and a subspace of features that are shared between them.
Because of these characteristics, we refer to this manifold as the \emph{partitioned subspace (PS) manifold}.
Finally, we provide several examples using both real and synthetic data to illustrate how the manifold can be applied in practice, and to establish intuition for its properties.

\textbf{Related work} 
The problem of learning parameters subject to subspace constraints has been investigated widely and for a variety of applications.
In general, these applications use only simple subspace constraints, which effectively cause the feasible set to be either the Grassmannian manifold or Stiefel manifold.
Optimization subject to general subspace constraints has been investigated much less.
\citet{kim2010line} proposed an algorithm for face recognition that is similar to our work in that they also learn a set of mutually-orthogonal subspaces.
However, their approach learns subspaces by incrementally updating and reorthogonalizing the subspace matrix, whereas we optimize on the constraint manifold itself using Riemannian optimization.
Our work presents, to our knowledge, the first formulation of general subspace constraints as a manifold.

The rest of this paper is organized as follows.
In section \ref{sec:background}, we describe background concepts and establish notation.
Next, we define the partitioned subspace manifold and provide details necessary to use it for optimization in section \ref{sec:psmanifold}.
Finally, section \ref{sec:experiments} illustrates the various properties of the manifold and analyzes them through experiments on both real and synthetic data.

\section{Background}
\label{sec:background}

Before defining the partitioned subspace manifold, we briefly introduce some necessary background concepts from differential geometry.

\subsection{Riemannian Manifolds} 
\label{subsec:riemannian_manifolds}

A \emph{Riemannian manifold}, or simply a manifold, can be described as a continuous set of points that appears locally Euclidean at every location.
More specifically, a manifold is a topological space and a collection of differentiable, one-to-one mappings called \emph{charts}.
At any point on a $d$-dimensional manifold, there is a chart that maps a neighborhood containing that point to the Euclidean space $\R{d}$ \cite{absil2009optimization}.
This property allows a \emph{tangent space} to be defined at every point, which is an Euclidean space consisting of directions that point along the manifold. 
Tangent spaces are particularly important for optimization because they characterize the set of update directions that can be followed without leaving the manifold.
While the most familiar manifold is Euclidean space, a large number of other, more exotic manifolds can be defined.
The manifolds discussed in this paper are examples of \emph{matrix manifolds}, meaning that they consist of points that can be represented as matrices. 

Given a manifold, it is often useful to define a \emph{quotient manifold} by specifying an equivalence relation that associates sets of points on the original manifold with individual points on the quotient \cite{edelman1998geometry}.
Because each point on the quotient is an equivalence class of points on the parent manifold, we use the notation $[p]$ to refer to a point on the quotient that contains $p$, where $p$ is an element of the parent manifold.
Quotient manifolds inherit a great deal of structure from their parent manifold.
In particular, constraints that define the parent manifold also apply to the quotient.

To illustrate this, consider the Stiefel and Grassmannian manifolds, which can be defined as quotients of the group of orthogonal matrices.
Later in section \ref{sec:psmanifold}, the derivation for the partitioned subspace manifold will follow a similar pattern.
The orthogonal group, denoted $\mathcal{O}_n$, consists of all $n \times n$ matrices $Q$ that satisfy the condition $Q^TQ = I$.
Note that in this work, we use $Q$ to refer to an element of this set.
From this definition, we see that each point $[Q]_{St}$ on a Stiefel manifold is a set of matrices from the orthogonal group whose first $k$ columns match those of $Q$.
Similarly, each point $[Q]_{Gr}$ on a Grassmannian manifold corresponds to those orthogonal group matrices whose first $k$ columns span the same subspace.
These relationships can be expressed mathematically as follows:
\begin{eqnarray*}
    \text{Stiefel points}\;[Q]_{St} &=& \left\{ Q E \right\}_{E \in \mathcal{E}_{St}} \\ 
    \text{Grassmannian points}\;[Q]_{Gr} &=& \left\{ Q E \right\}_{E \in \mathcal{E}_{Gr}}
\end{eqnarray*}
where
\begin{eqnarray*}
    \mathcal{E}_{St} &=& \left[ \begin{array}{cc} I_k & \\ & \mathcal{O}_{n-k} \end{array}\right], \;\;\text{and} \\
    \mathcal{E}_{Gr} &=& \left[ \begin{array}{cc} \mathcal{O}_k & \\ & \mathcal{O}_{n-k} \end{array}\right]
\end{eqnarray*}
Because they encode the relationship between matrices that map to the same point on the quotient, we refer to the sets $\mathcal{E}_{St}$ and $\mathcal{E}_{Gr}$ as the \emph{equivalence sets} for their respective manifolds.

\textbf{Representative elements}
As defined above, points on the Stiefel and Grassmannian manifolds are equivalence classes $[Q]$ of orthogonal matrices $Q \in \mathcal{O}_n$.
However, in practice, these points are represented by single, $n \times k$ matrices $Y$.
This discrepancy occurs because in general, only the subspace spanned by the first $k$ columns is of interest. 
Given a point $[Q]$ on these manifolds, we refer to $Y$ as a \emph{representative element} defined by:
\begin{equation*}
 Y = Q'\left[\begin{array}{c} I_k \\ 0\end{array}\right], \;s.t.\; Q' \in [Q]
\end{equation*}

\subsection{Riemannian Optimization}
\label{subsec:riemannian_optimization}

Optimization on manifolds requires specialized \emph{Riemannian optimization} methods to ensure that the learned parameters remain valid.
While many standard optimization methods have been derived on manifolds \cite{edelman1998geometry}, we focus on Riemmannian gradient descent because of its simplicity.
As in standard gradient descent, the goal is to use the gradient of a differentiable convex function $f$ with respect to the matrix $Q$, denoted $G_Q$, to construct a sequence of estimates $Q' = Q \; - \; \alpha G_Q$ that converges to the minimum of $f$, where the parameter $\alpha$ controls the size of the update at each step.
However, in Riemannian gradient descent, the estimates are constrained to lie on a manifold. 

\begin{figure}[t]
    \centering
    \vskip 0.1in
    \includegraphics[scale=0.32]{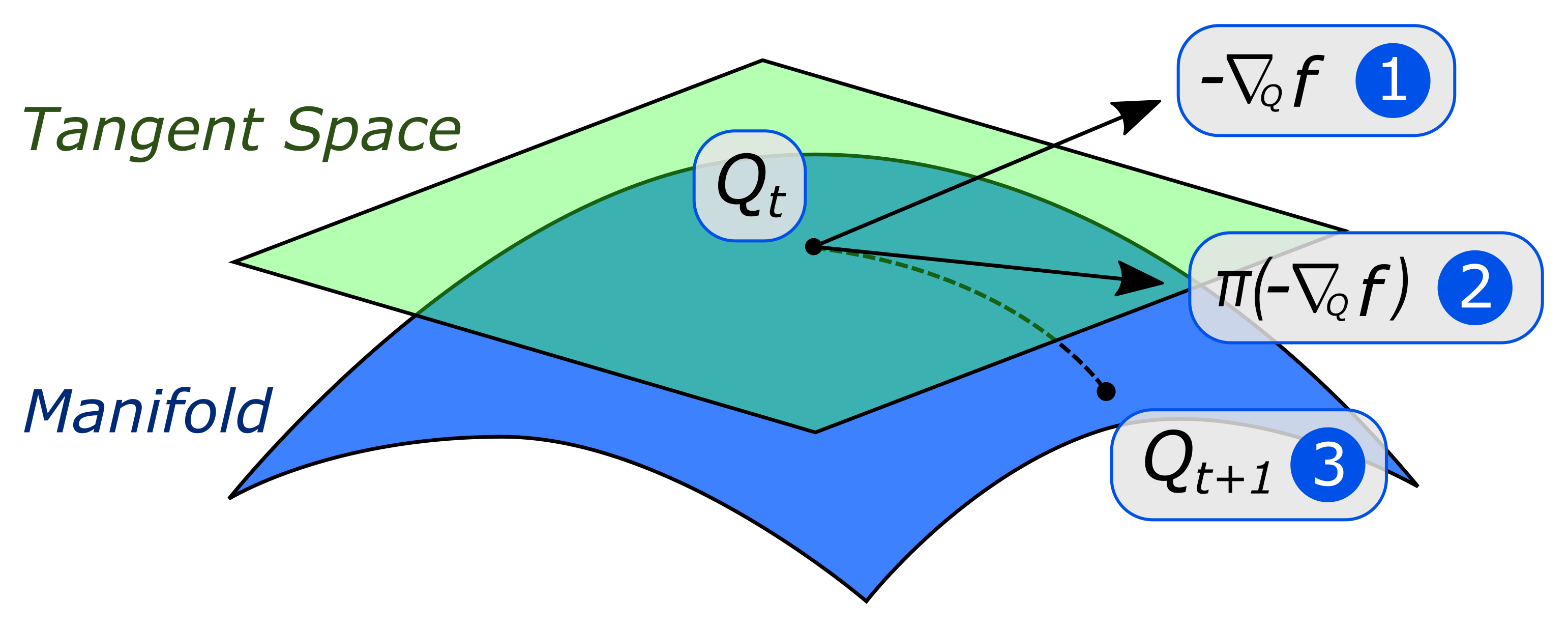}
    \caption{The three steps for Riemannian gradient descent. First, the negative gradient direction is computed. Next, that direction is projected onto the tangent space at the current estimate. Finally, a new estimate is computed using an appropriate retraction. These steps are repeated until the estimates converge. }
    \label{fig:rgd}
\end{figure}

Riemannian gradient descent proceeds by iterating three steps, as illustrated in Figure \ref{fig:rgd}.
The first step is to compute the gradient at the current location, which is given by $G_Q = \nabla_Q f$.

Next, the update direction is modified to reflect the geometry of the underlying manifold.
Intuitively, if some component of the update direction is orthogonal to the manifold's surface, then any movement at all along that direction will immediately leave the manifold.
To correct for this, the update direction is projected onto the tangent space at the current location. 
We denote the projected update direction as $\Delta = \pi_{T_QM}(G_Q)$, where $\pi_{T_QM}(\cdot)$ projects its input onto the tangent space of manifold $M$ at $Q$.

The final step of Riemmannian gradient descent computes a new location by moving along the manifold in the direction $\Delta$. 
This is accomplished using the \emph{exponential map}, which maps elements of a tangent space onto the manifold by moving in a straight line along its surface.
For matrix manifolds, the exponential map is given by the matrix exponential function $\exp(\cdot)$, and the updated location is given by the following formula, as shown in \cite{edelman1998geometry}:
\begin{equation}
Q' = Q\exp(\alpha Q^T\Delta)
\end{equation}
Unfortunately, computing updates in this way is inefficient due to the cost of evaluating the matrix exponential. 
Instead, it is common to use a \emph{retraction}, which behaves similarly to the exponential map but may distort the magnitude of the update. 
Depending on the manifold, these alternatives may be comparably inexpensive.
For example, updates on quotient manifolds of the orthogonal group can be performed using the \emph{Q-factor retraction}, which maps an update $\Delta$ at the point $Q$ to the location $Q' = \text{qr}(Q+\Delta)$, where $\text{qr}(\cdot)$ denotes the Q matrix of the QR decomposition of its input \cite{absil2009optimization}. 

The steps of computing the standard gradient, projecting onto the tangent space at the current point $Q$, and updating the location using an appropriate retraction are repeated until convergence. 

\section{The Partitioned Subspace Manifold}
\label{sec:psmanifold}

Having defined the necessary basic concepts, we now describe the partitioned subspace (PS) manifold, emphasizing the intuition behind its derivation. 
We also derive the components needed to apply Riemannian optimization on the manifold in section \ref{subsec:psm_optimization}.
Note that in this section, we use $[Q]$ to denote $[Q]_{PS}$ for notational convenience.

\subsection{Formulation}

The PS manifold is defined as a quotient of the orthogonal group, similarly to how the Stiefel and Grassmannian manifolds were defined in section \ref{subsec:riemannian_manifolds}.
Here, the difference will be the use of a generalized equivalence relation.
For the Stiefel manifold, the equivalence relation equates two orthogonal group matrices if they are related by a rotation of their last $n-k$ columns.
The Grassmannian equivalence relation, on the other hand, equates two matrices if they are related by both a rotation of their last $n-k$ columns and a rotation of their first $k$ columns.
The PS manifold equivalence relation generalizes this pattern by equating two matrices if they are related by rotation of the first $k_1$ columns, or the next $k_2$ columns, etc.
Intuitively, rotations that only involve basis vectors for a single partition will not change the subspace they span, whereas rotations that mix basis vectors from different partitions will.
Based on this, we define the partitioned subspace manifold as follows.

\begin{defn}
Let $k_1$, $k_2$, ..., $k_m$ be subspace partition sizes, where $m$ is the number of partitions.
Furthermore, let $n$ be the dimensionality of the ambient space, and let $k = \sum_{i=1}^mk_i \leq n$ denote the total subspace size.
We define the partitioned subspace manifold as a quotient of the group of orthogonal matrices, with points on the manifold characterized by:
\begin{equation}
    \text{PS points}\;[Q]_{PS} = \left\{ Q E \right\}_{E \in \mathcal{E}_{PS}}
\end{equation}
where
\begin{equation}
{\arraycolsep=2.6pt
    \mathcal{E}_{PS} = \left[ \begin{array}{cccc} \mathcal{O}_{k_1} & & & \\ & \ddots & & \\ & & \mathcal{O}_{k_m} & \\ & & & \mathcal{O}_{n-k}\end{array}\right]}
\end{equation}
\end{defn}
Similarly to the Stiefel and Grassmannian manifolds, the last $n-k$ columns are generally ignored, and points on the PS manifold are represented by $n \times k$ matrices $Y = Q'\left[\begin{array}{cc} I_k & 0\end{array}\right]^T$ for some $Q' \in [Q]_{PS}$.
It is straightforward to show that the partitioned subspace manifold is equivalent to a Grassmannian manifold when $m=1$ and $k_1=k$, and equivalent to the Stiefel manifold when $m=k$ and $k_j=1$ for $j=1,2,...,m$. 

Next, we characterize the tangent space for the PS manifold, which will be required for Riemannian optimization.
Let $Q$ be any element of an equivalence class defining a point on the PS manifold.
Because the manifold is a quotient of $\mathcal{O}_n$, it is subject to the orthogonality contraint $Q^TQ=I$.
As shown by \citet{edelman1998geometry}, this implies that elements $\Delta$ of the tangent space at $[Q]$ satisfy the skew-symmetry condition, $Q^T\Delta = -\Delta^TQ$.
It is straightforward to verify that $\Delta = QA$ satisfies this condition for all skew-symmetric $A$.
However, this does not fully characterize the tangent space at $[Q]$ because there are directions of this form that lead only to points that are in the same equivalence class $[Q]$.
These directions are exactly the ones for which movement in that direction is the same as multiplication by an element of $\mathcal{E}_{PS}$:
\begin{equation*}
Q\exp(\alpha Q^T\Delta) = QE, \;\; E \in \mathcal{E}_{PS}
\end{equation*}
Solving this equation reveals that updating $Q$ in the direction $\Delta$ will not change its equivalence class on the PS manifold if $\Delta$ is of the form:
\begin{equation*}
{\arraycolsep=2.6pt
    \Delta \;=\; Q\left[ \begin{array}{cccc} A_{k_1} & & & \\ & \ddots & & \\ & & A_{k_m} & \\ & & & A_{n-k}\end{array}\right],}
\end{equation*}
where each $A_i$ is a $k_i \times k_i$ skew-symmetric matrix. 
Note that to reach this conclusion, we rely on the observation that $Q^T\Delta = A$, and that $\exp(A) \in \mathcal{O}$ because $A$ is skew-symmetric \cite{edelman1998geometry}.
Given these observations, tangent spaces on the partitioned subspace manifold are characterized as follows.
Let $B_{i,j}$ denote an arbitrary $k_i \times k_j$ matrix, and let $B_{\perp,j}$ and $B_{i,\perp}$ denote arbitary $(n-k) \times k_j$ and $k_i \times (n-k)$ matrices, respectively.
The tangent space at a point $[Q]$ on the partitioned subspace manifold is the set of all $n \times n$ matrices $\Delta$ given by:
\begin{equation} \label{eq:ps_ts}
{\arraycolsep=2.6pt
     \Delta \;=\; Q\left[ \begin{array}{ccccc} 0         & -B_{2,1}^T & \cdots & -B_{m,1}^T & -B_{\perp,1}^T \\ 
                                           B_{2,1}   & 0          & \cdots & -B_{m,2}^T & -B_{\perp,2}^T \\ 
                                           \vdots    & \vdots     & \ddots & \vdots     & \vdots       \\ 
                                           B_{m,1}   & B_{m,2}    & \cdots & 0          & -B_{\perp,m}^T \\ 
                                           B_{\perp,1} & B_{\perp,2}  & \cdots & B_{\perp,m}  & 0             \end{array}\right]
}
\end{equation}

\subsection{Optimization on the PS Manifold} 
\label{subsec:psm_optimization}

To apply Riemannian optimization on the partitioned subspace manifold, we must derive the tangent space projection operator and identify an appropriate retraction.

The projection of an arbitrary update direction $Z$ onto the tangent space at $[Q]$ can be derived as follows.
First, we project $Z$ onto the set of matrices of the form $QA$, where $A$ is skew-symmetric.
The result is given by $Q\;\text{skew}(Q^TZ)$, where $\text{skew}(X) = \frac{1}{2}(X-X^T)$ \cite{edelman1998geometry}.
Next, we replace the diagonal blocks of this matrix with zeros to obtain the projection of $Z$ onto matrices of the form given in equation \ref{eq:ps_ts}.

Although deriving expressions for the $B_{i,j}$ blocks in equation \ref{eq:ps_ts} is tedious due to notation, it is straightforward and simplifies to the following. 
Let $Z_i$ and $Q_i$ denote the $k_i$ columns of $Z$ and $Q$ corresponding to the $j$th subspace partition.
For an arbitrary matrix $Z$, the projection onto the tangent space of the PS manifold at a point $Q$ is given by
\begin{equation} \label{eq:ps_tsproj_1}
    \pi_{T_QPS}(Z) = \frac{1}{2}\left[ \;\pi_1\; | \;\pi_2\; | \;...\; | \;\pi_m\; | \;\pi_{n-k}\; \right],
\end{equation}
where  
\begin{equation} \label{eq:ps_tsproj_2}
    \pi_i = (Z_i-QZ^TQ_i) + (Q_iZ_i^TQ_i - Q_iQ_i^TZ_i).
\end{equation}

With the tangent space projection defined, we now consider retractions on the partitioned subspace manifold.
Ideally the exponential map would be used, but the computational cost associated with it makes it impractical.
Fortunately, because our manifold is a quotient of the group of orthogonal matrices, the \emph{Q-factor retraction} can be applied to efficiently compute updates.

Having derived the tangent space projection and an appropriate retraction, Riemannian gradient descent can be applied as described in section \ref{sec:psmanifold}.
Our approach is summarized in Algorithm \ref{alg:pd-rgd}.


\begin{algorithm}[tb]
   \caption{Riemannian gradient descent on the partitioned subspace manifold}
   \begin{algorithmic}
      \STATE {\bfseries Input:} loss function $f(Q)$, \hspace{0.01in} initial estimate $Q_0$, \\ \hspace{0.375in} partition sizes $\{k_i\}_{i=1}^m$, \hspace{0.01in} step size $\alpha$
      \FOR{$i=0,1,2,...\;$}
      \STATE compute $G_Q \leftarrow \nabla_Q f(Q_i)$
      \STATE compute $\Delta \leftarrow \pi_{T_{Q_i}PS}(G_Q)$  \hspace{1cm}(Eqns.\ref{eq:ps_tsproj_1}-\ref{eq:ps_tsproj_2})
      \STATE compute $Q_{i+1} \leftarrow \text{qr}(Q_i - \alpha \Delta)$
      \IF{$Q_i \approx Q_{i+1}$}
      \STATE return $Q_{i+1}$ 
      \ENDIF
      \ENDFOR
   \end{algorithmic}
   \label{alg:pd-rgd}
\end{algorithm}

\section{Applications}
\label{sec:experiments}
To illustrate how the PS manifold can be used in practice, we applied it to the problems of multiple dataset analysis and domain adaptation.
Specifically, our intention throughout these applications was not to establish new state-of-the-art methods, but rather to provide proof of concept and to demonstrate the properties of the PS manifold.

\subsection{Multiple Dataset Analysis}

First, we use the PS manifold to extract features that describe a collection of datasets, grouped according to whether they are unique to one dataset in particular (the \emph{per-dataset features}) or shared between several (the \emph{shared features}).
Given a collection of $D$ datasets $\mathcal{X} = \{X_i\}_{i=1}^D$, we define $D+1$ partitions, where the first $D$ partitions will capture the per-dataset features and the last partition will capture the shared features.
We note that in general, additional partitions could be defined to model more complex relationships, e.g., to learn features unique to distinct pairs or triplets of datasets.
For simplicity, we consider only per-dataset and shared features in this section, and leave the problem of learning more general partitioning schemes as a promising direction for future work.

\subsubsection{Problem Formulation}

Our approach is motivated by the \emph{principal components analysis (PCA)}, which computes the $k$-dimensional subspace that best reconstructs a single input dataset \cite{dunteman1989principal}.
The subspace returned by PCA can be viewed as the solution to the following optimization problem:
\begin{equation*}
    Y_{PCA} = \underset{Y \in Gr(n,k)}{\arg\min}\;||X-XYY^T||^2_F\:,
\end{equation*}
where $\text{Gr}(n,k)$ denotes the Grassmannian manifold of $k$-dimensional subspaces of $\R{n}$, and $||\cdot||_F$ denotes the Frobenius norm.
The expression $||X-XYY^T||^2_F$ calculates the reconstruction error between the original dataset $X$ and its projection onto $Y$, which is given by $XYY^T$.
Based on this optimization problem, we propose \emph{Multiple-Dataset PCA} to solve the following related problem:
\begin{defn}
Let $X_i$ denote one of $D$ datasets in a collection $\mathcal{X}$, and let $PS(n,k_{pd},k_{sh})$ be a partitioned subspace manifold consisting of $D$ per-dataset partitions of size $k_{pd}$ and one shared partition of size $k_{sh}$. 
Furthermore, given a point $[Q]$ on the PS manifold, let $Q_i$ and $Q_{sh}$ denote the columns of $Q$ that span the $i$th per-dataset partition and the shared partition, respectively.
Finally, define $S_i = [\;Q_i\;|\;Q_{sh}\;]$ to be the union of the per-dataset partition for $X_i$ and the shared partition.
We define the Multi-Dataset PCA subspace to be the point on the PS manifold given by $Y_{\textit{MD-PCA}} = Q_{\textit{MD-PCA}} \left[\begin{array}{cc} I_k & 0\end{array}\right]^T$, where $Q_{\textit{MD-PCA}}$ is equal to:
\begin{equation} \label{mdpca}
    Q_{\textit{MD-PCA}} = \underset{Q \in PS(n,k_{pd}, k_{sh})}{\arg\min} \;\sum_{i=1}^D \frac{||X_i-X_iS_iS_i^T||^2_F}{|X_i|}. 
\end{equation}
\end{defn}

Given Algorithm \ref{alg:pd-rgd} for performing Riemannian gradient descent on the partitioned subspace manifold, the MD-PCA algorithm is simple to state: apply Riemannian gradient descent using the gradient of the loss function in Equation \ref{mdpca} at each step.
For MD-PCA, the gradient can be computed using the equations below.
\begin{eqnarray*}
    \frac{\partial}{\partial Q_i}f(Q) &=& -2\frac{X_i^TX_iQ_i}{|X_i|} \;\;\text{and} \\
    \frac{\partial}{\partial Q_{sh}}f(Q) &=& -2\left( \sum_{i=1}^D \frac{X_i^TX_i}{|X_i|} \right) Q_{sh}
\end{eqnarray*}

\subsubsection{Experiments}
\label{subsec:mdpca-exps}

To evaluate MD-PCA, we used The Office+Caltech10 set, which is a standard object recognition benchmark containing four datasets of processed image data \cite{gong2012geodesic}. 
Each dataset is drawn from a different source, which is either \emph{Amazon} (A), \emph{DSLR} (D), \emph{Caltech} (C) or \emph{Webcam} (W). 
The original images are encoded as 800-bin histograms of SURF features that have been normalized and z-scored to have zero mean and unit standard deviation in each dimension, as described in \cite{gong2012geodesic}. 
Each image is associated with one of ten class labels, but these were not used in our experiments because MD-PCA, like PCA, is an unsupervised learning approach.


\begin{figure}[t]
    \centering
    \includegraphics[scale=0.32]{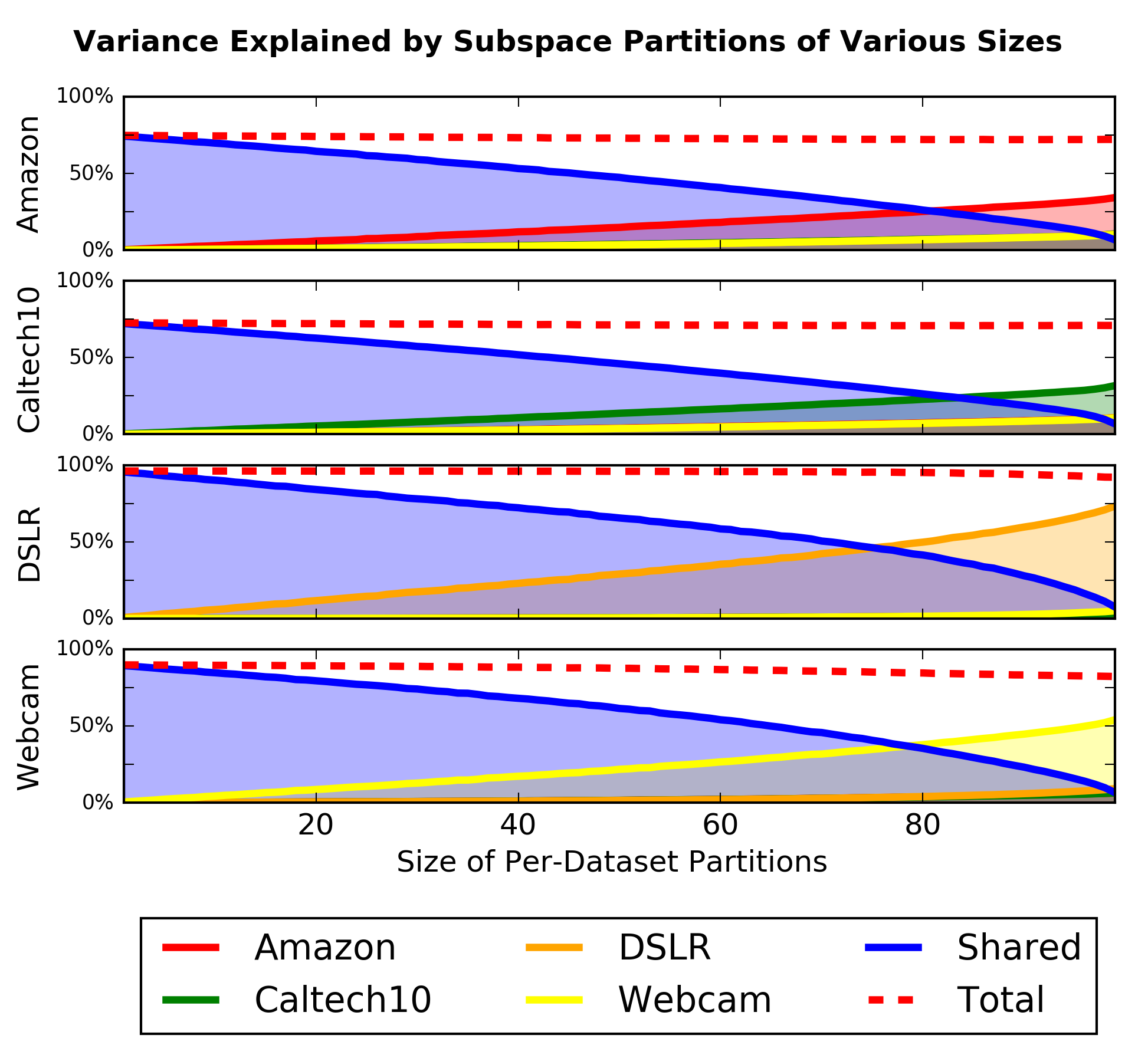}
    \caption{Results of MD-PCA for various sizes of $k_{pd}$. For each trial, $k_{sh}$ was set to $400-4k_{pd}$ so that the total number of components learned was fixed at $400$. For each dataset, the majority of its variance is explained by the shared partition and its per-dataset partition. Note that the total variance explained is relatively insensitive to the per-dataset partition sizes, suggesting that $k_{pd}$ can be set liberally in practice. }
    \label{fig:mdpcafigasdf}
\end{figure}

To evaluate the sensitivity of the MD-PCA results with respect to the chosen partition sizes, we performed an exhaustive sweep over values for $k_{pd}$.
Specifically, $k_{pd}$ was varied between $1$ and $99$, and the size of the shared partition was set to $400-4k_{pd}$.
Figure \ref{fig:mdpcafigasdf} illustrates the result of this analysis.

\subsubsection{Remarks}

MD-PCA highlights several of the benefits and considerations of using the PS manifold in practice.

First, the mutual orthogonality constraints between the subspace partitions guarantee that a feature in one partition is not contained in any other, i.e., the dataset-specific features truly are unique.
Because the manifold enforces mutual-orthogonality exactly, there is never redundancy in the features that might otherwise be introduced due to the constraints only being satisfied approximately.

Second, these same constraints cause common features to be included in $Q_{sh}$, and unique features to be included in the corresponding subspace $Q_i$, despite the lack of explicit discriminative terms in the loss function.
Intuitively, if a feature is important for reconstructing $X_i$, then the $i$th term in Equation \ref{mdpca} will encourage it to be contained in either $Q_i$ or $Q_{sh}$.
If that feature is also important for reconstructing a different dataset $X_j$, then the $j$th term will behave similarly, and the pressure to include it in $Q_{sh}$ will dominate the pressure to include it in $Q_i$ or $Q_j$.
On the other hand, if the feature is unique to $X_i$, then the loss terms for the other datasets will encourage $Q_{sh}$ to be populated with other features, so that the pressure to include it in $Q_i$ will be uncontested while the pressure to include it in $Q_{sh}$ will be neutralized.
This behavior arises because the mutual orthogonality constraints captured by the PS manifold prevent a feature from being in multiple partitions simultaneously.
If the partitions were not constrained in this way, for example if they were optimized as points on separate Grassmannian manifolds, then shared features would tend to mix with the per-dataset features, introducing redundancy and increasing the overall reconstruction error.

Finally, the use of the PS manifold simplifies the presentation of the algorithm by eliminating the steps that would otherwise have to be taken to enforce the mutual orthogonality constraints.

On the other hand, there are limitations to this approach.
Foremost, while MD-PCA extracts and partitions features for multiple datasets, those datasets must represent the data in a consistent way.
In particular, if samples from two datasets have different dimensionality, as is often the case with image datasets for example, then MD-PCA cannot be applied without first processing the datasets to map them to a common feature space.

Also, MD-PCA requires the user to specify not only the total subspace size, as in PCA, but also the size of each partition.
Figure \ref{fig:mdpcafigasdf} shows how the variance explained by the subspace partitions changes as the per-dataset partition size is varied. 
In general, the total variance explained for each dataset does not change much as the partition size changes, until the per-dataset partitions become large enough to cause the shared partition to be very small.
This observation implies that in practice, the size of the per-dataset partitions can be set liberally as long as the shared partition remains reasonably large.
We hypothesize that the decrease in explained variance occurs when the shared partition becomes too small to contain all of the common features, which causes one dataset to "steal" the feature into its partition while the reconstruction error suffers for the other datasets.

\subsection{Domain Adaptation}

\begin{table*}[t]
\caption{Domain adaptation results with \texttt{Amazon} and \texttt{Caltech10} as source domains.}
\label{table1-1}
\begin{center}
\begin{sc}
\begin{tabular}{l|ccc|ccc} 
\hline
\emph{DA Method} & A $\rightarrow$ C & A $\rightarrow$ D & A $\rightarrow$ W & C $\rightarrow$ A & C $\rightarrow$ D & C $\rightarrow$ W  \\ 
\hline
OrigFeat      & 22.6         & 22.2       & 23.5          & 20.8        & 22.0          & 19.4       \\ 
GFK(PCA, PCA) & 35.6        & 34.9         & 34.4         & 36.9        & 35.2          & 33.9       \\ 
GFK(PLS, PCA) & 37.9       & 35.2        & 35.7        & 40.4         & 41.1       & 35.8         \\ 
SA-SVM & \textbf{39.9}       & \textbf{38.8}        & 39.6        & 46.1         & 39.4       & 38.9         \\ 
\hline
PS-NB          &   37.7      &   36.3      &   \textbf{40.0}          &     \textbf{46.4}   &   \textbf{43.4}      &   \textbf{40.3}    \\ 
PS-SVM          &    35.1       &  31.4      &    24.2          &     41.7   &   40.9      &   29.3 \\ 
\hline
\end{tabular}
\end{sc}
\end{center}
\vskip -0.1in
\end{table*}

\begin{table*}[t]
\caption{Domain adaptation results with \texttt{DSLR} and \texttt{Webcam} as source domains.}
\label{table1-2}
\begin{center}
\begin{sc}
\begin{tabular}{l|ccc|ccc} 
\hline
\emph{DA Method} & D $\rightarrow$ A & D $\rightarrow$ C & D $\rightarrow$ W & W $\rightarrow$ A & W $\rightarrow$ C & W $\rightarrow$ D  \\ 
\hline
OrigFeat      & 27.7          & 24.8         &  53.1          &  20.7         & 16.1           & 37.3       \\
GFK(PCA, PCA) & 32.6         &  30.1         & 74.9         & 31.3        &  27.3           &  70.7         \\ 
GFK(PLS, PCA) & 36.2         & 32.7        & 79.1        &  35.5         & 29.3         &  71.2          \\ 

SA-SVM & \textbf{42.0}       & 35.0        & \textbf{82.3}        & \textbf{39.3}         & 31.8       & 77.9         \\ 
\hline

PS-NB          &   38.0         &     \textbf{35.8}      & 81.0                  &     39.2     &   \textbf{31.8}    &   \textbf{78.1}  \\ 
PS-SVM          &    32.0        &  31.8      &   53.8          &     30.6   &   23.2      &   77.9      \\           
\hline
\end{tabular}
\end{sc}
\end{center}
\vskip -0.1in
\end{table*}

To further illustrate using the partitioned subspace manifold, we applied it to the problem of domain adaptation.
Here, the goal is to leverage a labeled dataset $X_s$ sampled from a \emph{source distribution}, to improve accuracy when predicting an unlabeled target dataset $X_t$, sampled from a different \emph{target distribution}.
In this work, we consider the problem of performing classification in the target domain. 
Typical strategies for domain adaptation include projecting $X_s$ and $X_t$ onto a common space on which a classifier is then trained \cite{fernando2013unsupervised,gong2012geodesic,gopalan2011domain}, or finding a transformation of the datasets such that a classifier generalizes well to both domains \cite{Kulis:2011:YSY:2191740.2191798,ICML2012Chen_416}. 

The unique properties of the partitioned subspace manifold make it well suited to learning representations that generalize between the source and target.
In the following, we provide an illustrative example of using the PS manifold in this way.

\subsubsection{Problem Formulation}

In our approach, we wish to learn a set of subspace partitions that captures the common features between the source and target domains, while providing a representation that makes it easy to distinguish between classes.
To accomplish this, we hypothesize that there are class-specific subspaces on which the source and target distributions are equivalent, so that a classifier trained on the projected data will generalize well across domains.
Secondly, we constrain the subspaces to be mutually orthogonal to promote better discrimination between classes after projection.
The partitioned subspace manifold makes it straightforward to specify and implement this objective.

We note that while the strategy of identifying domain- or label-specific features is not new to domain adaptation, the use of explicitly partitioned subspaces to capture those features has, to the best of our knowledge, not been investigated. 

First, we define $Q$ to be a partitioned subspace, where each partition will contain features that describe source samples with a given label $y$. 
To encourage the subspace to capture the structure of the source and target domains, we minimize the reconstruction error of approximating $X_s$ and $X_t$ by their projections onto $Q$.
In addition, we promote discriminative subspaces by maximizing the magnitude of each source sample when projected onto its corresponding class partition, and minimizing the magnitude when projected onto the remaining partitions.
Combining these, we define the \emph{Class Discriminative Transfer (CDT) subspace} as follows. 

\begin{defn}
Let $X_s$ be a source dataset with labels $Y_s$, where $y_x$ denotes the label for sample $x$, and let $X_t$ be a target dataset.
Let $L$ denote total number of unique labels.
For each label $y_i$, we define a per-class partition $Q_{y_i}$ of size $k_{pc}$ to contain features that describe source samples with that label.
Let $Q = [\;Q_{y_1} \;|\; Q_{y_2} \;|\; \dots \;|\; Q_{y_L} \;|\; Q_\perp\;]$ be an orthonormal matrix containing these partitions in its first $k$ columns, where $k = k_{pc}L$ is the combined size of the partitions.
Finally, let $Q_{\bar{y}}$ be the union of the class-specific partitions that do not correspond to label $y$.
We define the Class Discriminative Transfer (CDT) subspace to be $Y_{CDT} = Q_{\textit{CDT}} \left[\begin{array}{cc} I_k & 0\end{array}\right]^T$, where $Q_{\textit{CDT}}$ is equal to:
\begin{equation}
    Q_{CDT} = \;\underset{\mathclap{Q \in PS(n,k_{pc})}}{\arg\min} \;\;f(Q),
\end{equation}
and
\begin{equation}
{\arraycolsep=1.4pt\def\arraystretch{1.6}
\begin{array}{rl}\label{eq:cst-loss}
f(Q) =& ||X_s - X_sQQ^T||^2_F + ||X_t - X_tQQ^T||^2_F \\
      & \;\;\; - \lambda \underset{\mathclap{x \in X_s}}{\sum} \;\Big(||xQ_{y_x}Q_{y_x}^T||^2_F - ||zQ_{\bar{y}_x} Q_{\bar{y}_x}^T||^2_F\Big)
\end{array}}
\end{equation}
\end{defn}

The first two terms in Equation \ref{eq:cst-loss} minimize the reconstruction error of the projected source and target data, while the third term encourages subspaces that allow samples from different classes to be easily distinguished.
As a result, a set of partitions is learned that accurately reconstructs the source and target datasets while grouping the features into partitions that are useful for classification.

\subsubsection{Experiments}

We evaluated our approach using the Office+Caltech10 dataset described in Subsection \ref{subsec:mdpca-exps}.
Unlike the multiple dataset analysis application, here we use the provided labels to conduct supervised training on the source domain and to evaluate test accuracy on the target domain.
Experiments were conducted for every possible pair of source and target domains, leading to $12$ configurations.

To measure the effectiveness of using the CST subspace for domain adaptation, we compared against two commonly used approaches that are related to ours.
The first, called the \emph{Geodesic Flow Kernel (GFK)} method, computes one subspace to represent the source data and another to represent the target.
A similarity between samples is then computed by integrating their inner product after projection onto all intermediate subspaces between the source and target. 
The similarities are then used as input to a 1-NN classifier \cite{gong2012geodesic}.
We include results for two versions of GFK: in the configuration GFK(PCA,PCA), the source and target subspaces are learned using PCA, while in GFK(PLS,PCA), partial least squares (PLS) is used to learn a source subspace that captures covariance between the source data and the available labels \cite{wold2001pls}.

The second approach we compare to is \emph{Subspace Alignment (SA)}, which first computes source and target subspaces in a similar way to GFK.
However, SA then projects each dataset onto its respective subspace and determines the optimal linear transformation to align the projected source data to the projected target data.
After the transformation is applied, a support vector machine classifier is trained \cite{fernando2013unsupervised}.
Finally, we compared our approach to the simple baseline of training an SVM on the original SIFT features without applying domain adaptation, which we call \emph{OrigFeat} in our results.
To ensure a fair comparison, we used the same evaluation protocols used for GFK and SA for our experiments \cite{gong2012geodesic,fernando2013unsupervised}. 

For our experiments, we conducted trials using partition sizes $k_{cs}$ of 20 and 80 dimensions. 
Note that because the Office+Caltech10 datasets consist of samples with $800$ features across $10$ classes, the maximum class-specific partition size is 80. 
In all cases, the 80-dimensional subspace partitions performed best.   
The value of $\lambda$ was fixed to be $2.0$ for all experiments.
After learning the CST subspace, we evaluated classification accuracy using Support Vector Machines (SVM) and Naive Bayes (NB). 
Tables \ref{table1-1} and \ref{table1-2} show the mean accuracy on all transfer tasks. 

\subsubsection{Remarks}

When comparing accuracy using SVM as a classifier, the partitioned subspace manifold is able to achieve comparable performance to GFK and SA. 
This suggests that the different partitions are able to learn discriminative subspaces that generalize to the target domain. 
Our approach also seems to be particularly well suited to using Naive Bayes for classification. 
In this case, the predictions are not made according to distance between the data, but according to the probability distributions of each feature.

\section{Conclusion}

In this work, we presented a formulation for the \emph{Partitioned Subspace Manifold}, which captures the geometry of mutually orthogonal subspaces, and provided the necessary components for optimizing on the manifold using Riemannian optimization techniques.
The ability to specify and easily optimize under general subspace constraints offers a significant amount of flexibility in designing new loss functions. 
To illustrate this, we proposed and analyzed multiple-dataset PCA, an extension of the principle component analysis which is able to group features from multiple datasets according to whether they are unique to a specific dataset or shared between them.
In addition, we showed how the PS manifold can be used to learn class-specific features for domain adaptation and proposed a method that achieves comparable accuracy several standard approaches. 
We are actively interested in extending this formulation to allow subspace partitions to be adaptively resized so that they can be more effective in online settings, where the optimal partition sizes may change. 

\bibliography{bibliography}
\bibliographystyle{aaai}

\end{document}